\documentclass{article}

\PassOptionsToPackage{numbers}{natbib}

\usepackage[preprint]{neurips_2021}

\usepackage[utf8]{inputenc} 
\usepackage[T1]{fontenc}    
\usepackage{hyperref}       
\usepackage{url}            
\usepackage{booktabs}       
\usepackage{amsfonts}       
\usepackage{nicefrac}       
\usepackage{microtype}      
\usepackage{xcolor}         
\RequirePackage{tabularx}
\usepackage{graphicx}
\usepackage{epsfig}
\usepackage{dcolumn}
\usepackage{subcaption,soul}
\newcommand{\etal}{\textit{et al}. }
\usepackage{amssymb}

\usepackage{threeparttable}
\usepackage{multirow}
\usepackage{subcaption}
\usepackage{float}

\title{MMRNet: Improving Reliability for Multimodal Object Detection and Segmentation for Bin Picking via Multimodal Redundancy}

\author{Yuhao Chen\textsuperscript{1}, Hayden Gunraj\textsuperscript{1}, E. Zhixuan Zeng\textsuperscript{1}, \\
\textbf{Robbie Meyer\textsuperscript{1}, Maximilian Gilles\textsuperscript{2}, Alexander Wong\textsuperscript{1}}\\
 \textsuperscript{1} University of Waterloo, Canada\\
 \textsuperscript{2} Karlsruhe Institute of Technology, Germany\\
{\tt\small \{yuhao.chen1, hayden.gunraj, ezzeng, robbie.meyer, alexander.wong\}@uwaterloo.ca},\\
{\tt\small maximilian.gilles@kit.edu}
}

\begin{document}

\maketitle

\begin{abstract}
Recently, there has been tremendous interest in industry 4.0 infrastructure to address labor shortages in global supply chains. 
Deploying artificial intelligence-enabled robotic bin picking systems in real world has become particularly important for reducing stress and physical demands of workers while increasing speed and efficiency of warehouses.
To this end, artificial intelligence-enabled robotic bin picking systems may be used to automate order picking, but with the risk of causing expensive damage during an abnormal event such as sensor failure.
As such, reliability becomes a critical factor for translating artificial intelligence research to real world applications and products. 
In this paper, we propose a reliable object detection and segmentation system with \textbf{M}ulti\textbf{M}odal \textbf{R}edundancy (MMRNet) for tackling object detection and segmentation for robotic bin picking using data from different modalities. 
This is the first system that introduces the concept of multimodal redundancy to address sensor failure issues during deployment.
In particular, we realize the multimodal redundancy framework with a gate fusion module and dynamic ensemble learning.
Finally, we present a new label-free multimodal consistency (MC) score that utilizes the output from all modalities to measure the overall system output reliability and uncertainty. 
Through experiments, we demonstrate that in an event of missing modality, our system provides a much more reliable performance compared to baseline models.
We also demonstrate that our MC score is a more reliability indicator for outputs during inference time compared to the model generated confidence scores that are often over-confident.
\end{abstract}

\begin{figure}[th!]
    \centering
    \includegraphics[width=0.5\columnwidth]{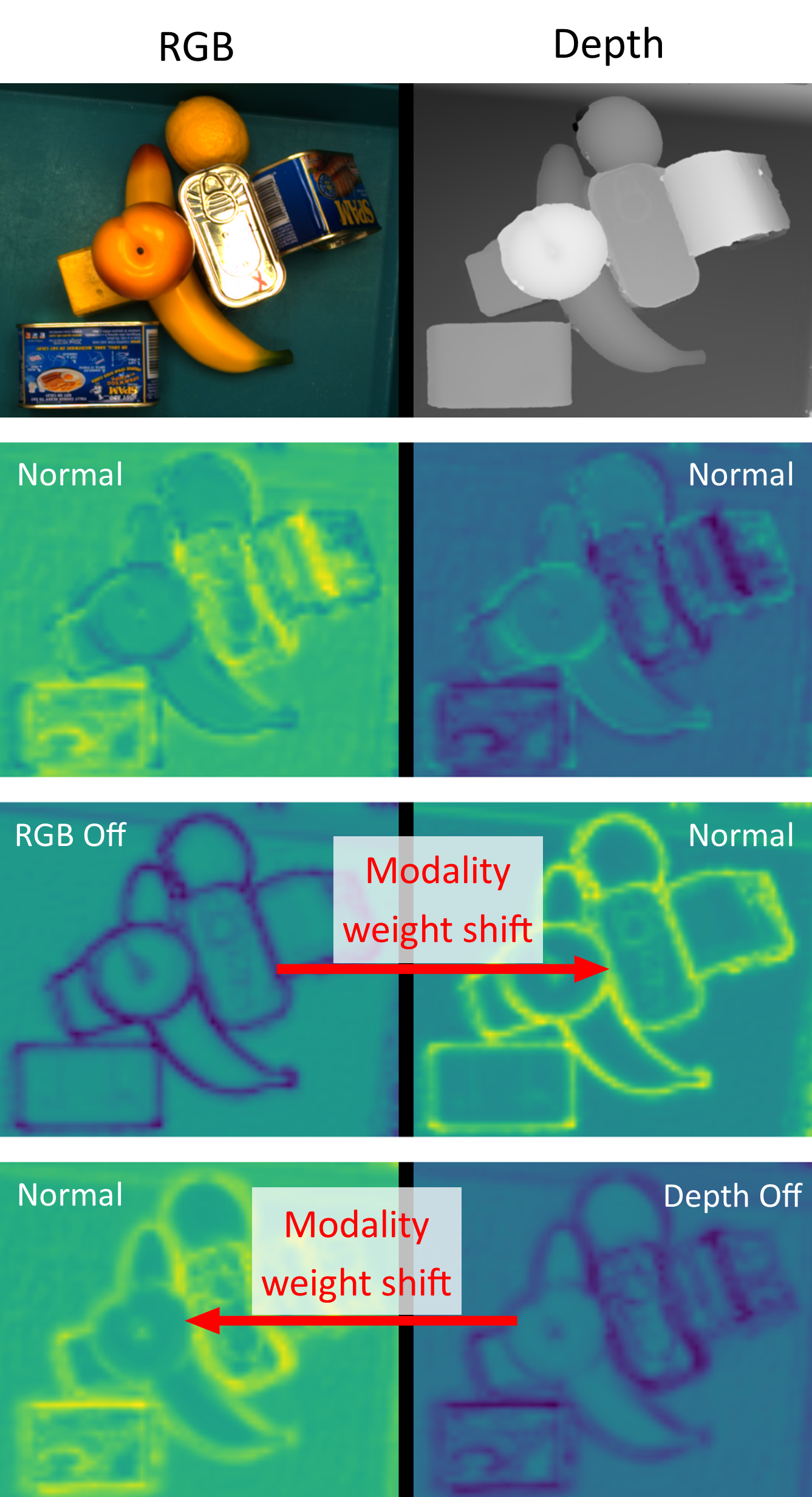}
    \caption{The dynamic modality weight shifting of our network ensures a reliable overall performance when a modality is missing.
    Row 2-4 heatmaps describe the average gate weights of each modality at a single feature scale. Yellow indicates high weight, dark purple indicates low weight.
    }
    \label{fig:gates}
\end{figure}
\section{Introduction}
Global labor shortages and the need for resilient supply chains has accelerated companies' upgrades to industry 4.0 and introduced a range of technologies such as big data, cloud computing, internet of things (IoT), robotics, and artificial intelligence (AI) into production systems.
With warehouses and manufacturing units becoming smart environments,  a crucial objective is to develop an autonomous flow of both material and information, and robotic bin picking plays an essential role in this task.

Robotic bin picking has been an active area of research for many decades given the complexity of the task, ranging from joint control and trajectory planning \cite{ichnowski2022gomp} to object identification \cite{uoais} and grasp detection \cite{suctionnet1b}.
In particular, we examine the object detection and segmentation task in autonomous bin picking. 
Different from object detection and segmentation in other areas such as autonomous driving, robotic vision system works in environments that are very close to the camera, dealing with heavy occlusions, shadows, dense object layouts, and complex stacking relations.
It plays an essential role in a robot's perception system.

Deep neural networks have been proven effective for object detection and segmentation \cite{maskrcnn, yolo}. But, deploying such systems in robotic picking applications is challenging due to the many sources of uncertainty present in practical scenarios. Real-world bin scenes may consist of a wide variety of unknown or occluded items arranged in an infinite number of poses and illuminated with variable lighting conditions. 
In addition to the variability of real-world bin scenes, errors in the camera system can make a computer vision system unreliable. Camera sensors are prone to noise and can fail in various situations such as specular reflections (missing values), black areas (missing depth), overexposure, blur, and artifacts. In practice, commercial systems are expected to run 24/7 to be feasible, which increases the risk of imaging sensor failures compared to research environments. If not accounted for, sensor failures can lead to wrongly commissioned orders and in the worst case to product and hardware damages, leading to expensive recall campaigns or production downtime. Therefore, vision systems capable of handling uncertain inputs and producing reliable predictions under sensor errors are critical to creating fail-safe applications.


One approach to creating fault-tolerant object detection and segmentation systems is to introduce system duplication, where portions of the system are duplicated to allow the system to continue to operate despite failures of its constituent parts. 
This approach assumes that failures are caused by either input sensor failures or computational failures. 
However, duplication may not provide fault-tolerance in situations where the system is operating correctly but its sensors are unable to adequately measure the inputs. 
For example, a camera may fail to adequately image a piece of glass due to its transparency, and so the use of a second identical camera cannot address this issue. 
In addition, deep neural networks as a data-driven approach are designed to capture feature distributions of the input dataset.
A simple duplication of these networks will not detect features that are not in the training distribution.
Instead, we add image data from depth sensor as an additional modality to capture object feature characteristics from a different perspective. 
More specifically, depth data has very simple texture yet rich geometric features, that are more transferable to unseen objects than RGB data.

A good system duplication design duplicates components that are more likely to fail, preventing any disruption in the information flow from the system input to output.
Non-data-driven methods have well defined explicit logic to control the information flow.
In comparison, deep learning system learns the input and output mapping through high-dimensional implicit feature representations.
A typical deep learning model encodes input information through a backbone network into a high-dimensional latent representation, and downstream tasks use the representation to predict low-dimensional outputs.
Consequently, a large amount of information is lost during the dimensionality reduction of downstream tasks.
However, in robotic bin-picking, unseen items may contain highly complex image characteristics that require both RGB and depth to work collaboratively. 
For example, RGB backbones are better at detecting transparent objects and depth backbones are better at detecting dark objects. 
A pair of eyeglasses with black frame will require the RGB backbone to focus on glass parts while the depth backbone to focus on the frame for a complete detection and segmentation.
With the reduction of dimensionality, a simple result aggregation on two low quality detections will create another low quality detection. 
Additional result merging networks or explicit merging logic will introduce errors and instabilities into the system.
An effective modality fusion technique that will dynamically fuse modality features with limited loss of information is therefore greatly desired.
In addition, modality features merging may introduce dependencies between them, causing unexpected model behavior when one of the modality feature is absent.
We tackle this problem with a multimodal redundancy framework consists of two key techniques: 
1) we use a multi-scale soft-gating mechanism to make the network learn to weigh and combine features between modalities dynamically, and
2) we use a dynamic ensemble learning strategy to train the sub-system independently and collaboratively in an alternating fashion.
With this framework, only one modality needs to be present for the model to operate.

Finally, we propose a novel multimodal consistency (MC) score as a more objective reliability indicator for the system output based on the overlaps of detected bounding boxes and segmentation masks. This can be used as an indicator for model uncertainty on individual predictions, as well as model reliability on particular datasets.

Through experiments, we demonstrate that in an event of missing modality, our MMRNet provides a much more reliable performance compared to baseline models.
When depth is removed, our network's performance drop is within 1\% where other models have a performance drop greater than 6\%.
When RGB is removed, our network's performance drop is within 11\% where other models have a performance drop greater than 80\%. 
Furthermore, we demonstrate that our MC score is a more reliable indicator for output confidence during inference compared to the often overly-confident confidence scores.
We summarize our contribution as the following:
\begin{itemize}

    \item A multimodal redundancy framework consisting of a multi-scale soft-gating feature fusion module and a dynamic ensemble learning strategy allowing trained sub-systems to operate both independently and collaboratively.
    \item A multimodal consistency score to describe the reliability of the system output.
    
\end{itemize}

\section{Related Work}
\textbf{Reliability study for deep learning-based systems:} 
Deep learning-based methods are data-driven, encoding the decision making process through continuous latent vectors, which makes the model behavior hard to predict and fix. 
Only a few of the studies focus on the reliability aspect of the deep learning-based systems.
In \cite{reliable_dl_systems}, Santhanam \etal list differences between traditional and deep learning-based software systems and discuss the challenges involved in the development of reliable deep learning-based systems.
In \cite{driving_reliability}, Xu \etal study the reliability of object detection systems in autonomous driving. 
In \cite{reduced_precisin_reliability}, dos Santos \etal study the relationship between reliability and GPU precision (half, single, and double) for object detection tasks.
Other reliability related work can be found in model uncertainty estimation \cite{uncertainty_survey}.
To the best of our knowledge, none of the work investigates reliability or uncertainty for multimodal applications, in particular for robotic bin picking.

\textbf{Multimodal Data Fusion:} Multimodal learning \cite{multimodal_deep_learning, xai_multimodal_review, multimodal_survey, multimodal_obj_det, multimodal_better} has been rigorously studied.
In multimodal learning, there are three types of data fusion: early fusion, intermediate fusion, and late fusion. 
Each corresponds to merging information at input, intermediate, and output stage respectively. 
Early fusion involves combining and pre-processing inputs.
A simple example is replacing the blue channel of RGB with depth channel \cite{early_fusion_rgd}. 
Late fusion merges the low-dimensional output of all networks. 
For example, Simonyan \etal \cite{late_fusion_video} combine spatial and temporal network output with i) averaging, and ii) linear Support Vector Machine \cite{svm}.
Early fusion and late fusion are simpler to implement but have a lower dimensional representation compared to the intermediate fusion.
Intermediate fusion involves merging high-dimensional feature vectors. 
Common intermediate fusion includes concatenation \cite{multimodal_deep_learning}, and weighted summation \cite{uoais}.
Recently, more advanced techniques are developed to dynamically merge the modalities. 
In \cite{wang2020cen}, Wang \etal propose a feature channel exchange technique based on Batch Normalization's \cite{batch_normalization} scaling factor to dynamically fuse the modalities.
In \cite{cao2021}, Cao \etal propose to replace the basic convolution operator with Shapeconv to achieve RGB and depth fusion at the basic operator level.
In \cite{hard_gate_fusion}, Xue \etal focus on the efficiency aspect of multimodal learning and propose a hard gating function which outputs an one-hot encoded vector to select modalities.
In robotic grasping, Back \etal \cite{uoais} take the weighted summation approach and propose a multi-scale feature fusion module by applying a 1x1 convolutional layer to the feature layers before passing them into a feature pyramid network (FPN)~\cite{feature_pyramid_network}. 

The aforementioned works are designed to optimize the overall network performance but at the same time introduce dependencies among modality features, which are extremely vulnerable in case of an abnormal event, such as an input sensor failure.
In this paper, we address the multimodal fusion strategy from the system reliability perspective, where our goal is to design a simple yet effective network architecture that enables sub-modal systems to work independently as well as collaboratively to increase the overall system reliability.

\textbf{Ensemble learning:} Ensemble learning typically involves training multiple weak learners and aggregating their predictions to improve predictive performance~\cite{watermelon}. One of the simplest approaches to construct ensembles is bagging~\cite{bagging}, where weak learners are trained on randomly-sampled subsets of a dataset and subsequently have their predictions combined via averaging or voting techniques~\cite{watermelon}. Instead of aggregating predictions directly, one may also use a meta-learner which considers the input data as well as each weak learner's predictions in order to make a final prediction, a technique known as stacking~\cite{stacking}. Boosting~\cite{adaboost} is another common approach where weak learners are added sequentially and leverage the previous learner's mistakes to re-weight training samples, effectively attempting to correct the previous learner's mistakes.

While ensemble learning has long been a common technique in classical machine learning, it can be expensive to apply to deep learning due to the increased computational complexity and training time of deep neural networks. Of particular relevance to this work is the application of ensemble learning to multimodal deep learning problems. In multimodal problems, the data distributions typically differ significantly between modalities and thus may violate the assumptions of certain ensembling techniques~\cite{marinoni2021}. Nevertheless, ensemble methods have been applied to a variety of multimodal problems~\cite{marinoni2021,menon2021,chordia2020,zhou2021}. For example, Menon~\etal~\cite{menon2021} trained modality-specific convolutional neural networks on three different magnetic resonance imaging modalities and combined the models' predictions via majority voting. In~\cite{zhou2021}, Zhou~\etal used a stacking-based approach to combine the outputs of neural networks trained on text, audio, and video inputs, thereby reducing noise and inter-modality conflicts.

Rather than combining multiple models with a typical ensembling strategy, in this work we consider a \textit{dynamic} ensemble where multiple unimodal systems are dynamically fused into a single network. This network is capable of both unimodal operation using each of its inputs independently as well as multimodal operation through the fusion of the constituent unimodal systems.

\section{Methodology}
  \begin{figure*}[t]
    \centering
    \includegraphics[width=1\columnwidth]{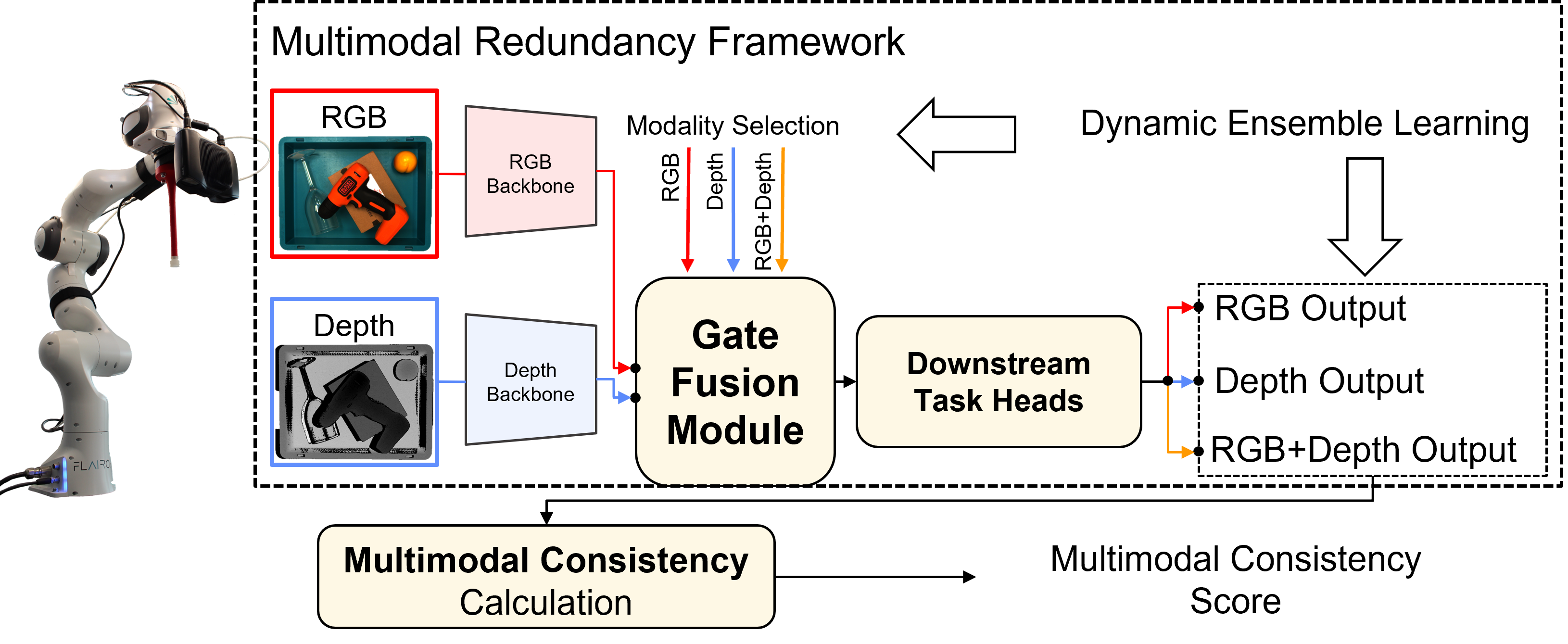}
    \caption{Block diagram of our multimodal redundancy framework. 
    Gate fusion module allows simple switching between modalities. Trained with dynamic ensemble learning, our system is able to use both modalities independently (RGB or depth output) as well as collaboratively (RGB+depth output). A multimodal consistency score is computed at the end to indicate the reliability of the output.}
     \label{fig:block}
   \end{figure*}
The subsequent sub-sections outline the key components of our MMRNet architecture. Firstly, we introduce a multi-scale soft gating mechanism that effectively combines information from the two modalities. Secondly, we propose a dynamic ensemble learning strategy, which, in conjunction with the multi-scale soft gating mechanism, constitutes the multimodal redundancy framework. This framework helps to remove the inter-modality dependencies.
Lastly, we present the formulation of the multimodal consistency score, which serves as our system's reliability measure.
We show our system block diagram in Figure \ref{fig:block}.

\subsection{Multimodal Redundancy Framework}
  \begin{figure}[t]
    \centering
    \includegraphics[width=0.6\columnwidth]{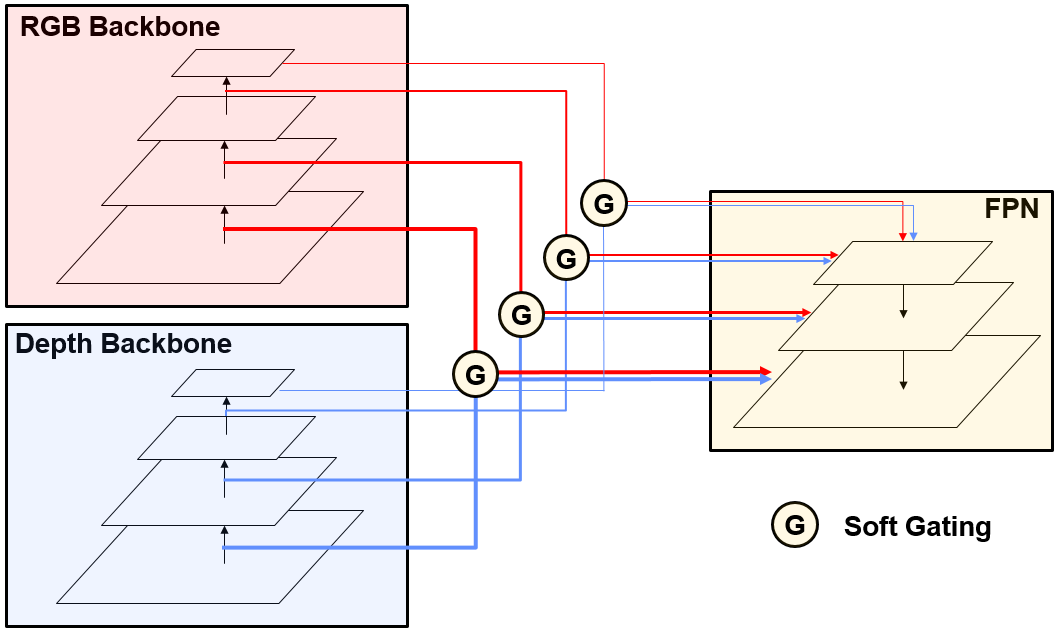}
    \caption{Gate fusion module fuses the multi-scale feature from each modality.}
     \label{fig:dense}
    \vspace{-3mm}
   \end{figure}
  \begin{figure}[t]
    \centering
    \includegraphics[width=0.6\columnwidth]{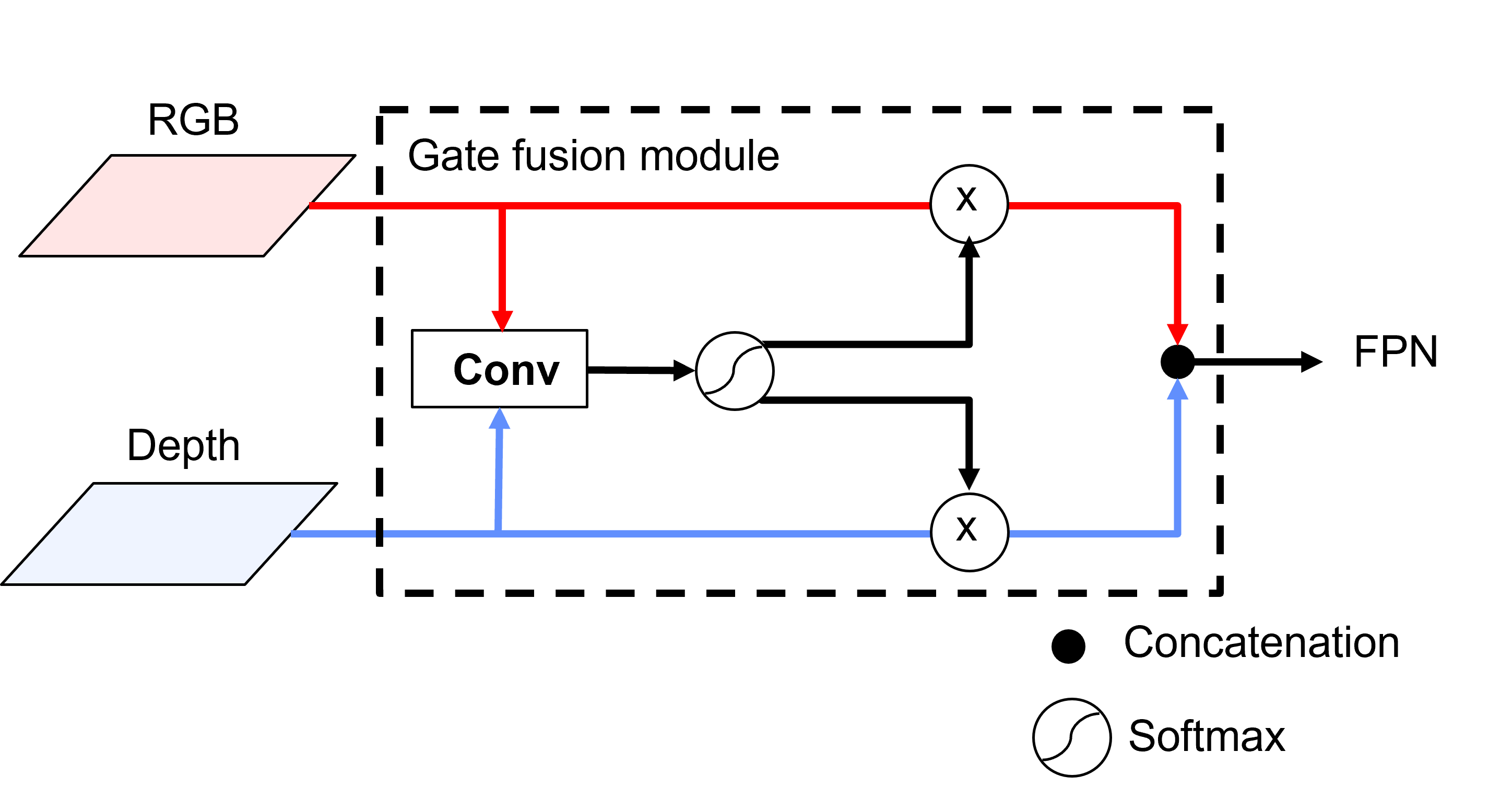}
    \caption{Soft gating architecture applied to every scale of feature layers.} 
     \label{fig:gating}
    \vspace{-4mm}
   \end{figure}
\textbf{Multi-scale Soft-Gate Feature Fusion (MSG Fusion): }
Fusing high-dimensional latent representation from two data distribution involves integrating information from multiple scales as well as multiple modalities. 
While a simple convolution as proposed in \cite{uoais} can merge the information, it also constrains the information exchange between modalities to be within the same scale.
The other modality's high-level features may contain crucial contextual information for localizing and segmenting objects with intricate RGB and depth features. In order to maximize the utilization of contextual information from both modalities, we concatenate the features and input them into a Feature Pyramid Network (FPN)~\cite{feature_pyramid_network}. This FPN fuses multi-scale modality features in a hierarchical manner, enabling effective contextualization.
Nonetheless, this process can result in inter-modality dependencies. To address this issue, we draw inspiration from \cite{soft_gate} and incorporate a soft gating mechanism. This mechanism enables the dynamic adjustment of feature weights from each backbone, thereby facilitating modality feature selection that is optimized for detecting individual object classes.
More importantly, this method enables the model to disentangle features from different modality backbones.
We define the total number of modalities to be $N$ and denote the $j$th scale feature layer in modality $m$ as $f_{m, j}$.
Features of all modalities pass through a 1x1 convolution layer $G_m$.
The convolution layer takes $N$ $j$th scale modality features with $C$ channels and outputs one feature layer with $C$ channels for modality $m$.
We obtain $g_{m, j}$:
\begin{equation}
    g_{m, j} = G_m \left( \left\{ f_{m, j}| m \in \left[ 0, N \right) \right\} \right)
\end{equation}
The output gate weight $w_{m, j}$ is calculated by:
\begin{equation}
    w_{m, j} = \sigma(\left\{ g_{m, j}| m \in \left[ 0, N \right) \right\})
\end{equation}
, where $\sigma$ is the softmax function ensuring modality weights sum to one.
Finally, the gated feature layer for scale $j$ and modality $m$ is updated by:
\begin{equation}
    f_{m, j} \leftarrow f_{m, j}w_{m, j}
\end{equation}
We show the gate fusion module architecture in Figure~\ref{fig:dense} and Figure~\ref{fig:gating}.

\textbf{Dynamic Ensemble Learning Strategy:} Although the proposed soft gating mechanism enables dynamic re-weighting of features extracted from each input modality, it does not inherently allow for modalities to be used independently. Ideally, the network would be capable of operating reasonably using a single input modality, with each additional modality providing improved performance or reliability. This accounts for the practical scenario where the sensor used to capture an input modality fails, forcing the system to leverage its other inputs. 

Classic ensemble approaches combine weak models according to their standalone performance by a simple discrete process such as weighted sum~\cite{watermelon}.
In comparison, our gating module allows more dynamic interaction since information can be exchanged across different modalities and scales with respect to different input items, poses, and scene layouts.
Instead of independently training each modality model and combine them with a weighted sum, we propose a novel dynamic ensemble learning strategy to train multimodal deep learning models, allowing for different modalities to be used both collaboratively and independently. 

Specifically, in each training iteration we randomly select one of the possible input conditions: both inputs, RGB-only input, or depth-only input. In the unimodal conditions, we force the system to make predictions with only one of its usual inputs in order to encourage rich features to be extracted from both modalities. This training scheme prevents the model from learning to rely heavily on a single modality while simultaneously allowing the model to learn how to combine data from both modalities.

\subsection{Multimodal Consistency Score}
Existing object detection and segmentation networks contain a confidence score calculated by the softmax classifier for each detection.
The reliability information in this score is somewhat subjective as it is estimated from the same network.
Instead, we leverage the multimodal property of our model.
In an ideal scenario, if we train a separate model for each modality, all models would converge to produce the same output describing the same object in the physical space. 
Less reliable models will produce results that deviated from the ground truth.
Models trained with different modalities capture distinct feature distributions and characteristics such as textures and geometries. 
We assume the output deviation between the modalities is very different from each other.
If the network output is reliable, then the outputs between modalities are well-aligned. This can be measured by the percentage overlap between output bounding boxes as well as the segmentation masks.
Based on this assumption, we argue that the more deviations between the modalities there are, the less certain the output is. 
To estimate the deviation, we use Intersection Over Union (IOU). 
It is a ratio between the intersection of the two modalities and their union and can be applied to boxes as well as masks.
Given a pair of detection/segmentation output $x_0$ and $x_1$. 
Each represents a set of pixels. 
$x_0$ and $x_1$ can either be a pair of boxes or a pair of masks.
Then, IOU can be calculated by:
\begin{equation}
    IOU(x_0, x_1) = \frac{|x_0 \cap x_1|}{|x_0 \cup x_1|}
\end{equation}
Where $|.|$ is a function that computes the number of pixels for the given input.
When deviation becomes larger, IOU will be smaller.
When there is less deviation, IOU will be larger and close to 1. 
This behavior captures well the output alignment between modalities.
When comparing results in object detection/segmentation, object matching is involved.
There can be multiple detections for one object, so we average the IOU score for all related detections associated with this object.
For simple annotation, we call the two models being compared source and target. 
Source results are matched to the target results.
Let the set of all $n_s$ detections in source be $D_s=\left\{d_{s, l}|l \in \left[ 0, n_s\right)\right\}$.
We compute the IOU of all items in $D_s$ associated with the $k$th target detection $d_{t, k}$, and obtain a set of IOUs $I_{D_s, d_{t, k}} = \left\{IOU(d_{s, i},d_{t, k} )|d_{s, i} \in D_s\right\}$. 
We define objects with IOU lower than 30\%, a typical threshold value used in the Non-Maximum Suppression (NMS) step in object detection networks such as \cite{maskrcnn}, as non-matched and remove them.
The updated IOU set is 
\begin{equation}
I^{'}_{D_s, d_{t, k}} = \left\{a | a \in I_{D_s, d_{t, k}}, a > 0.3 \right\}
\end{equation}
Next, we compute the average IOU $A(D_s, d_{t, k})$ for source detections $D_s$ and target detection $d_{t, k}$:
\begin{equation}
    A(D_s, d_{t, k}) = mean(I^{'}_{D_s, d_{t, k}})
\end{equation}
We further compute the mean IOU for all $n_t$ detections in target $D_t$:
\begin{equation}
    mIOU(D_s, D_t) = mean(\left\{A(D_s, d_{t, k}) | k \in [0, n_t)\right\})
\end{equation}
We extend this mIOU score to describe the alignment between our network output and all the modalities.
We name this score multimodal consistency (MC) score.
MC can be used to describe the alignment of one modality or multiple modalities in a multimodal system.
Let $D_o$ to be the network output with $n_o$ number of detections, and $D_m$ to be the network output using only modality $m$.
The MC score $S_{m}$ for a single modality $m$ is calculated by:
\begin{equation}
    S_{m} = mIOU(D_m, D_o)
\end{equation}
The MC score $S$ for all modalities is computed by:
\begin{equation}
    S = mean(\left\{A(D_m, d_{o, k}) | m \in [0, N), k \in [0, n_o)\right\})
\end{equation}
The higher the MC score is, the more reliable the system is.
A score of 100\% means all modalities predict the exact same output, and the system is very reliable.
A score of 0\% means each modality predicts a different output, and the system is unreliable.

\section{Experiments}

\subsection{Dataset and Implementation Details} 
Among robotic grasping datasets, the MetaGraspNet dataset~\cite{metagraspnet2022} provides large-scale, high-resolution simulated RGB and depth data as well as real-world data from an industry-grade sensor system.
In addition, the dataset contains 82 objects and has a novel object set for testing.
We divide the real dataset into train, validation, test, and test novel. 
We first exclude all scenes with novel objects, adding them to a separate novel test data split.
Then we split the rest of the real dataset into 80\% train, 10\% validation, and 10\% test. 

Due to the unique characteristics of each modality, we normalize and pre-process RGB inputs and depth inputs differently.
We use the standard mean variance normalization for RGB inputs and we apply min-max normalization per scene for depth inputs, where depth values are min-max normalized to $[0, 1]$.
We further flip the depth values to make 0 as the depth of the background and 1 as the closes point to the camera. 
With this value flip, background values are aligned to be 0 in each scene.
In addition, this normalization added a data augmentation to the dataset as it stretches and compress object shapes in depth, allowing a fully utilized depth range where every depth value is used by an object. 

Near objects' edges, reflective surfaces, and transparent surfaces, there are often undefined values caused by a lack of signal returning to the depth sensor. As a pre-processing step, we apply image inpainting \cite{inpaint_ns} to the depth images to replace any invalid values.

We use a classic object detection and segmentation network Mask-RCNN \cite{maskrcnn} with ResNet50 \cite{resnet} backbone as our baseline. 
All the networks in our experiment are initialized by the same ImageNet \cite{imagenet_comp} pretrained weights.
We train all models with the same training configuration in terms of batch size, training epoch, and optimizer. 
We pretrain all the models on the simulated dataset of MetaGraspNet, and finetune on the real dataset.
We report the performance of our method on the real test set with bounding box mean average precision (box mAP) and segmentation mask mean average precision (mask mAP).

\subsection{Results and Discussions}
\begin{table*}[tbp]

\begin{center}
\begin{threeparttable}

\begin{tabular*}{1\columnwidth}{@{\extracolsep{\fill}}lcccccc}

\toprule
        & & &  \multicolumn{2}{c}{Class agnostic} & \multicolumn{2}{c}{Class agnostic novel}\\
        \cmidrule{4-5}\cmidrule{6-7}
        Network & Box mAP & Mask mAP &  Box AP &  Mask AP &  Box AP &  Mask AP\\ 
\midrule        
        Baseline-RGB &  80.6 & 79.3 &  83.6 & 78.9 & 29.7 & 32.7\\ 
        Baseline-Depth &  73.1 & 71.9 &  77.2 & 72.0 & \textbf{32.3} & 32.1\\ 
        Baseline-Fusion\cite{uoais} &  80.3 & 79.4 &  82.8 & 78.8 & 28.7 & 33.1 \\ 
        MSG Fusion &  \textbf{81.7} & \textbf{80.7} &  \textbf{84.5} & \textbf{80.3} & 30.8 & \textbf{35.0}\\
\bottomrule        
\end{tabular*}
\scriptsize
\caption{Modality feature fusion comparison. Results (in \%) are compared to Mask R-CNN \cite{maskrcnn} unimodal baselines as well as multimodal feature fusion approaches.  }
\label{table:softgate-results}

\end{threeparttable}
\vspace{2mm}

\begin{threeparttable}

\begin{tabular*}{\columnwidth}{@{\extracolsep{\fill}}lcccccccc}
\toprule
 & \multicolumn{2}{c}{Box} & \multicolumn{2}{c}{Mask} & \multicolumn{2}{c}{RGB off} & \multicolumn{2}{c}{Depth off} \\ \cmidrule{2-3}
\cmidrule{4-5}\cmidrule{6-7} \cmidrule{8-9}   Network & AP       & MC   & AP       & MC   & Box AP   & Mask AP  & Box AP   & Mask AP  \\ \hline
Baseline-Fusion\cite{uoais}             & 82.8     & 43.0 & 78.8     & 41.0 & 0        & 0        &  75.2    & 68.6     \\

MSG Fusion                            & 84.5     & 39.1 &  \textbf{80.3}    & 36.7 & 0.6      & 0.3      & 78.1     & 74.3     \\
\quad + dynamic ensemble                & \textbf{84.9}     & \bf 82.9 & 80.2 &  \bf 84.7 & \bf 76.6 & \bf 69.6 & \bf 84.0 & \bf 79.3 \\ \hline
\end{tabular*}
\scriptsize
\caption{Class-agnostic results for multimodal architectures illustrating the need for dynamic ensemble learning for multimodal redundancy. Best results highlighted in \bf bold.}
\label{table:ensemble-results}
\end{threeparttable}

\end{center}
\end{table*}
\textbf{Multi-scale Soft-Gate Feature Fusion (MSG Fusion):} In Table \ref{table:softgate-results}, we show the improvement using our MSG Fusion under several train/test condition (regular class, class-agnostic, and novel objects). 
In general, we have observed that RGB data serves as a superior input modality for known objects. However, the performance of depth data tends to catch up in the detection of novel objects, as depth is more focused on object geometry.
RGB image data in general provides rich texture features that can be easily used to differentiate items, but may become a distraction for the network leading to false positives.
On the other hand, depth data provides simple geometric information that depicts object's shape with smooth surfaces.
Depth modality provides better performance when there is no RGB texture, dense texture creates distractions, and objects are too dark.
The performance improves even more when RGB and depth are both utilized.
The results show that dense fusion module provides better performance for novel items, while gated fusion strategy still provides competitive results.
The slight performance drop from dense fusion to gating mechanism can be due to the simplicity of the network without the gating module. 
The gating module limits the information flow, but enables the independent training of sub-systems.

\textbf{Dynamic Ensemble Learning Strategy for
Multimodal System:}
In this ablation study, we train our networks with and without the dynamic ensemble learning strategy and evaluate the networks' performance under three different input conditions: both RGB and depth, depth-only, and RGB-only. As shown in Table~\ref{table:ensemble-results}, the Gate Fusion network performs well after both standard learning and dynamic ensemble learning when both inputs are present. While dynamic ensemble learning yields slightly lower mask AP (-0.1 AP), it also results in greater box AP (+0.4 AP). More notably, when the RGB input is removed, the network trained without the dynamic ensemble strategy fails catastrophically (i.e., -83.9 box AP, -80.0 mask AP). This indicates that the network is over-reliant on RGB image information due to that fact that RGB images were always available during training. In contrast, the network trained with dynamic ensemble learning is able to mitigate the loss of RGB information by leveraging depth information alone. While the overall performance is markedly reduced in this scenario (as expected based on the Baseline-Depth results in Table~\ref{table:softgate-results}), it is a drastic improvement over the baseline Gate Fusion model. When depth inputs are turned off, the baseline model is still capable of reasonable operation but experiences a marked drop in performance. Adding dynamic ensemble learning mitigates this and results in only a minor loss of performance.

\textbf{Interpreting Gates:} The use of multi-scale soft gating offers a way to interpret the dynamic behaviour of a model under different input conditions. By averaging gate values over channels for a particular input scenario, we can obtain a human-interpretable heatmap for each input modality which illustrates the primary spatial regions being selected for by the gate. Figure~\ref{fig:gates} illustrates this technique for the soft gates at the second-highest spatial resolution of the feature pyramid. When both inputs are present, the soft gating mechanism primarily focuses on RGB features, although depth features are weighted strongly around the edges of the fruits (likely due to their homogeneity in RGB). However, once the RGB input is removed, the weights switch over to depth features dramatically in an attempt to compensate for the missing RGB features. Similarly, when depth features are removed, we see that RGB features are weighted more heavily around the edges of the fruits. This dynamic weight shifting between modalities is precisely the purpose of the soft gating mechanism and dynamic ensemble training strategy.

\begin{figure}
    \centering
    \begin{subfigure}[b]{0.45\linewidth}
         \centering
         \includegraphics[height=\textwidth, angle=90,trim={0 0.2cm 0 0.6cm},clip]{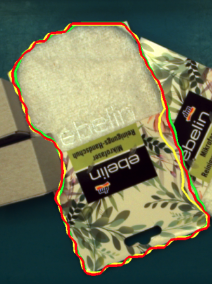}
         \caption{Train set object\\ MC score: 0.964 \\ Confidence score: 0.999}
         \label{fig:consis-train}
    \end{subfigure}
    \begin{subfigure}[b]{0.45\linewidth}
         \centering
         \includegraphics[height=0.97\textwidth, angle=270,trim={0.5cm 0 0.2cm 0},clip]{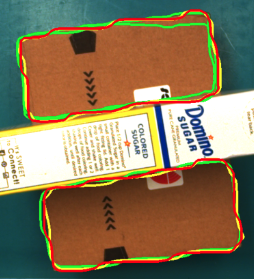}
         \caption{Test set object\\ MC score: 0.778 \\ Confidence score: 0.999}
         \label{fig:consis-test}
    \end{subfigure} \\
    \begin{subfigure}[b]{0.45\linewidth}
         \centering
         \includegraphics[height=\textwidth, angle=90,trim={0.2cm 0 0.2cm 0},clip]{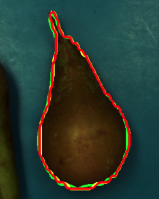}
         \caption{Test-novel set object\\ MC score: 0.966 \\ Confidence score: 0.999}
         \label{fig:consis-pear}
    \end{subfigure}
    \begin{subfigure}[b]{0.45\linewidth}
         \centering
         \scalebox{1}[-1]{\includegraphics[height=\textwidth, angle=90]{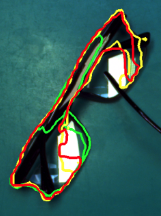}}
         \caption{Test-novel set object \\ MC score: 0.633 \\ Confidence score: 0.999}
         \label{fig:consis-glasses}
    \end{subfigure}
    \caption{Examples of object level MC scores. Gate fusion output is marked in red contour. RGB and depth only are marked in yellow and green contours respectively. In Figure \ref{fig:consis-test}, some predictions identified the object as separate boxes, decreasing the MC score.
    }
    \label{fig:consistency-examples}
    
\end{figure}

\textbf{MC Score:}
The model exhibits a steady decline in MC score between the training set, test set, and test-novel set as seen in Table \ref{table:class_agn_consistency}. This shows that the MC score decreases accordingly the more out-of-distribution a dataset is, correlating well with theoretically how reliable the model will be on each dataset.
The MC score also differs dramatically between objects of different classes. Objects with poor MC scores include disinfection bottle, glass bottle, cables in transparent bag, eyeglasses, and so forth, while boxes, cups, cables (not in plastic bags), and pears have higher MC scores, as seen in Table \ref{table:class_consistency}. This shows the ability of the MC score to identify challenging objects in the dataset.
We also compute the MC score against only RGB input or only depth input. Some objects exhibit a significant difference in MC score between those two options in Table \ref{table:class_consistency_modalities}. The starkest contrast appears when the object's material properties result in significant noise in one sensor, such as when transparent or reflective objects causes errors in the depth sensor. Through this, we can identify when the model is highly reliant on a particular sensor for its predictions. Eyeglasses, for example, performs both poorly overall (Table \ref{table:class_consistency}, Figure \ref{fig:consis-glasses}), and relies heavily on RGB input due to the transparency and reflection of its glass component.
\begin{table}[tbp]
\begin{center}
\begin{threeparttable}

\begin{tabular}{lcc}
\hline
Data Split                & MC Score - Box              & MC Score - Mask              \\ \hline
train                  & 91.4             & 92.7             \\
test & 82.9             & 84.7             \\
test-novel    & 73.4             & 73.7             \\ \hline
\end{tabular}
\vspace{2mm}
\caption{Class-agnostic MC scores from different data splits}

\label{table:class_agn_consistency}

\end{threeparttable}
\end{center}
\end{table}

\begin{table}[tbp]
\begin{center}
\begin{threeparttable}
\begin{tabular}{llll}
\hline
 &              & \multicolumn{2}{l}{MC Score (\%)} \\ \cline{3-4} 
 & Class        & Box               & Mask              \\ \hline
\multirow{4}{*}{\begin{tabular}[c]{@{}l@{}}non-novel\\ objects\end{tabular}} & disinfection bottle                                                          & 60.5 & 62.6 \\
 & glass bottle & 76.9             & 64.7             \\
 & cups d       & 96.0             & 96.5             \\ \hline
\multirow{4}{*}{\begin{tabular}[c]{@{}l@{}}novel \\ objects\end{tabular}}                     & \begin{tabular}[c]{@{}l@{}}cables in\\  \quad transparent bag\end{tabular} & 68.4 & 61.0 \\
 & eyeglasses      & 73.6             & 63.7             \\
 & pear         & 77.6             & 91.4            \\ \hline
\end{tabular}
\vspace{2mm}
\caption{MC scores for different object classes}
\label{table:class_consistency}
\end{threeparttable}
\end{center}
\end{table}

\begin{table}[tbp]

\begin{center}
\begin{threeparttable}

\begin{tabular}{lcc}
\hline
            & \multicolumn{2}{c}{Mask MC Score (\%)} \\ \cline{2-3} 
Class       & RGB only             & Depth only            \\ \hline
power drill & 68.4                & 73.3                 \\
wineglass   & 89.8                & 76.5                 \\
eyeglasses     & 69.6                & 58.0                 \\ \hline
\end{tabular}
\vspace{2mm}
\caption{Comparing MC score using only RGB or depth}

\label{table:class_consistency_modalities}
\end{threeparttable}
\end{center}
\end{table}

Examples of object-level MC scores for segmentation mask detections can be seen in in Figure \ref{fig:consistency-examples}. Note that the confidence score predictions for each object remains at an inflated 0.999 for all four examples, while the MC score shows a greater distinction between the objects depending on difficulty. This is especially true in Figure \ref{fig:consis-glasses}, where the model outputs a poor detection, but with high confidence. This, supported by previous dataset-level results, shows that the MC score is a better indicator for model reliability and uncertainty compared to the confidence score.

\section{Conclusion}

In this paper, we addressed the reliability aspect of deep learning-based computer vision systems for robotic grasping by introducing a multimodal redundancy framework called MMRNet. In particular, we realized the multimodal redundancy by two architecture changes, including a multi-scale dense modality feature fusion module and a gate fusion module to allow simple modality selection at per feature level. 
We also introduced a dynamic ensemble learning strategy to train modality models independently and collaboratively at the same time. 
Finally, we proposed a multimodal consistency score as a certainty indicator for the network output.
Results show that our MMRNet provides reliable and superior performance in the event of a modality input failure. 
We could also show that MC score is an well suited output reliability indicator independent of the network's confidence score.

\begin{ack}
We would like to thank the German Federal Ministry of Economic Affairs, the National Research Council Canada, Festo, and DarwinAI for supporting this work.
\end{ack}

\bibliographystyle{plainnat}  
\bibliography{references.bib}

\newpage
\appendix    
\section{Additional Results}
\begin{figure*}
    \centering
    \begin{subfigure}[b]{0.17\textwidth}
         \centering
         \includegraphics[width=\textwidth]{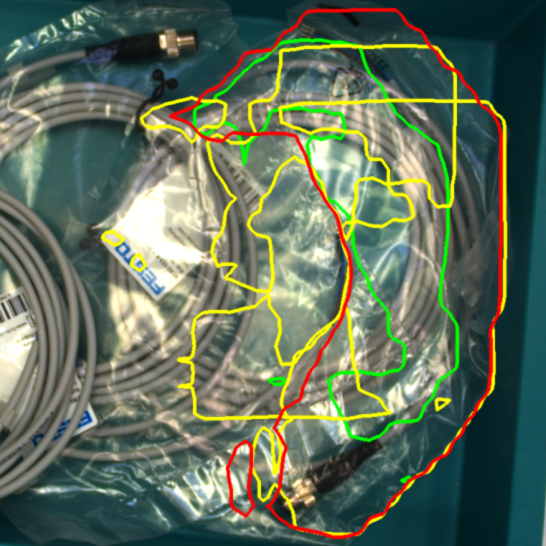}
         \caption{MC score: 0.538 \\ Confidence: 0.953}
         \label{fig:cables_in_transparent_bag}
    \end{subfigure}
    \begin{subfigure}[b]{0.17\textwidth}
         \centering
         \includegraphics[width=\textwidth]{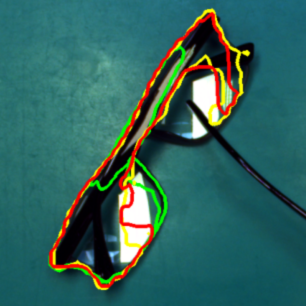}
         \caption{MC score: 0.633 \\ Confidence: 0.999}
         \label{fig:glasses}
    \end{subfigure} 
    \begin{subfigure}[b]{0.17\textwidth}
         \centering
         \includegraphics[width=\textwidth]{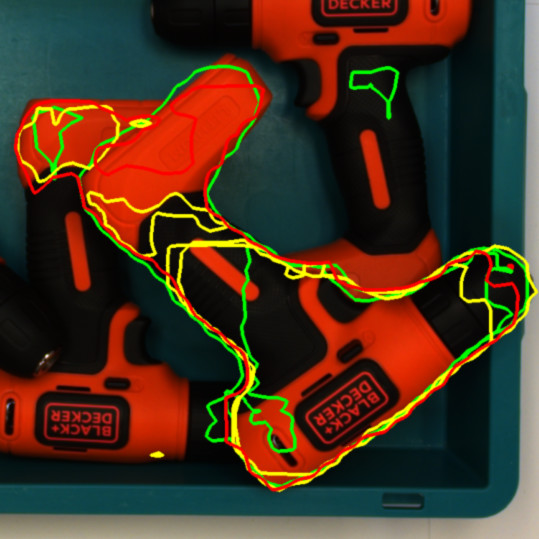}
         \caption{MC score: 0.669 \\ Confidence: 0.994}
         \label{fig:power_drill}
    \end{subfigure}
    \begin{subfigure}[b]{0.17\textwidth}
         \centering
         \includegraphics[width=\textwidth]{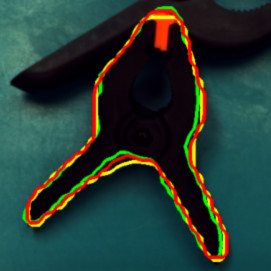}
         \caption{ MC score: 0.906 \\ Confidence: 0.999}
         \label{fig:clamp}
    \end{subfigure} 
    \begin{subfigure}[b]{0.17\textwidth}
         \centering
         \includegraphics[width=\textwidth]{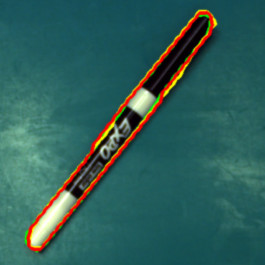}
         \caption{ MC score: 0.913 \\ Confidence: 0.999}
         \label{fig:marker_small}
    \end{subfigure} \\
    
    \begin{subfigure}[b]{0.17\textwidth}
         \centering
         \includegraphics[width=\textwidth]{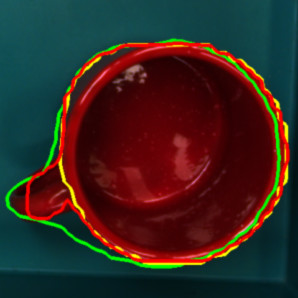}
         \caption{ MC score: 0.916 \\ Confidence: 0.999}
         \label{fig:mug}
    \end{subfigure} 
    \begin{subfigure}[b]{0.17\textwidth}
         \centering
         \includegraphics[width=\textwidth]{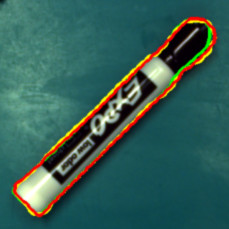}
         \caption{ MC score: 0.931 \\ Confidence: 0.999}
         \label{fig:marker_big}
    \end{subfigure}
    \begin{subfigure}[b]{0.17\textwidth}
         \centering
         \includegraphics[width=\textwidth]{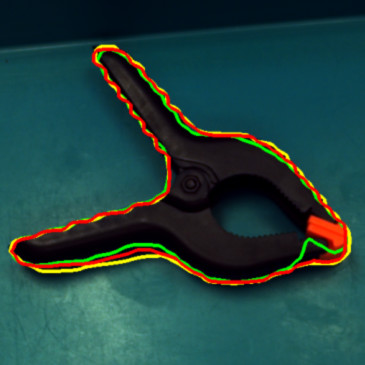}
         \caption{ MC score: 0.878 \\ Confidence: 0.999}
         \label{fig:clamp_big}
    \end{subfigure} 
    \begin{subfigure}[b]{0.17\textwidth}
         \centering
         \includegraphics[width=\textwidth]{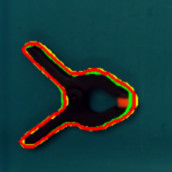}
         \caption{ MC score: 0.934 \\ Confidence: 0.999}
         \label{fig:clamp_small}
    \end{subfigure}
    \begin{subfigure}[b]{0.17\textwidth}
         \centering
         \includegraphics[width=\textwidth]{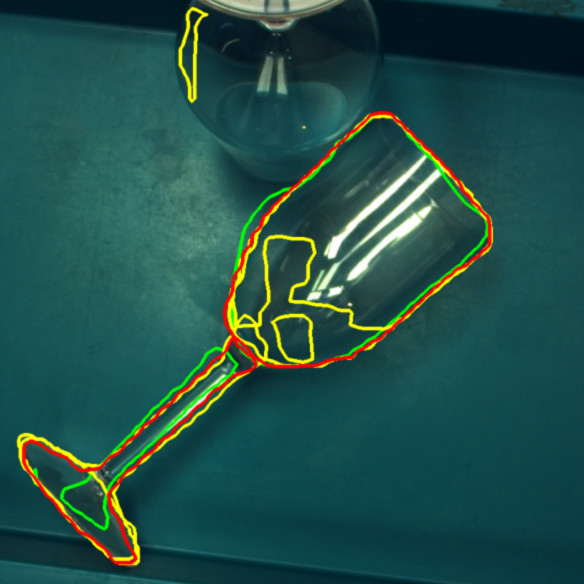}
         \caption{ MC score: 0.704 \\ Confidence: 0.991}
         \label{fig:wineglass}
    \end{subfigure} \\
    
    \begin{subfigure}[b]{0.17\textwidth}
         \centering
         \includegraphics[width=\textwidth]{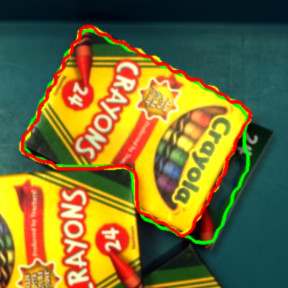}
         \caption{ MC score: 0.931 \\ Confidence: 0.999}
         \label{fig:crayons}
    \end{subfigure}
    \begin{subfigure}[b]{0.17\textwidth}
         \centering
         \includegraphics[width=\textwidth]{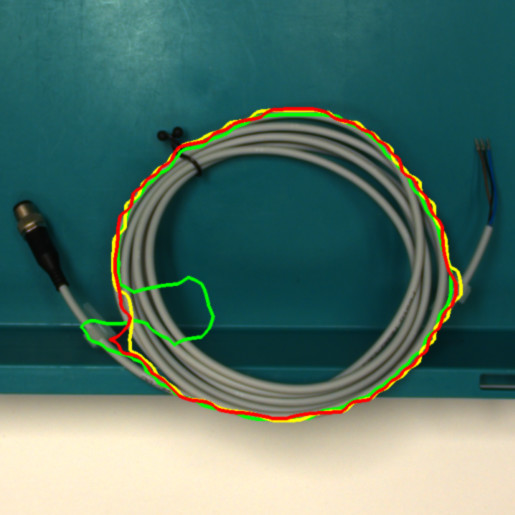}
         \caption{ MC score: 0.657 \\ Confidence: 0.999}
         \label{fig:cables}
    \end{subfigure}
    \begin{subfigure}[b]{0.17\textwidth}
         \centering
         \includegraphics[width=\textwidth]{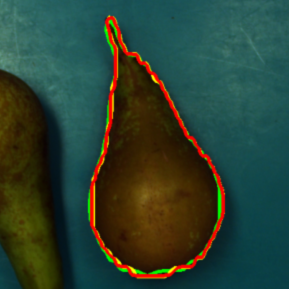}
         \caption{ MC score: 0.967 \\ Confidence: 0.999}
         \label{fig:pear}
    \end{subfigure}

    \caption{Image examples of objects from the novel-test set. The MC score shown is calculated at per object level instead of being averaged over the entire object category. We also show output confidence score for comparison. Gate fusion output is marked in red contour. RGB and depth only are marked in yellow and green contours respectively.
    }
    \label{fig:consistency-examples-novel}
    
\end{figure*}

We show complete Multimodal Consistency (MC) score results for our real-world test-novel experiment in Table \ref{table:novel_res_table}. 
This table contains results for all object categories in the test-novel set which was unable to fit into the page limit.
We also show randomly selected image results from this experiment for every object categories in test-novel dataset in Figure \ref{fig:consistency-examples-novel}.

\begin{figure*}
    \centering
    \begin{subfigure}[b]{0.17\textwidth}
         \centering
         \includegraphics[width=\textwidth]{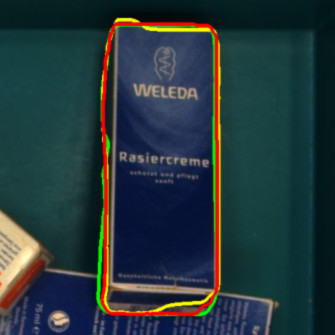}
         \caption{ MC score: 0.955 \\ Confidence: 0.999}
         \label{fig:desinfection}
    \end{subfigure}
    \begin{subfigure}[b]{0.17\textwidth}
         \centering
         \includegraphics[width=\textwidth]{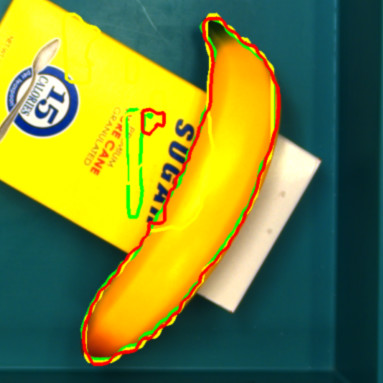}
         \caption{ MC score: 0.794 \\ Confidence: 0.980}
         \label{fig:banana}
    \end{subfigure}
    \begin{subfigure}[b]{0.17\textwidth}
         \centering
         \includegraphics[width=\textwidth]{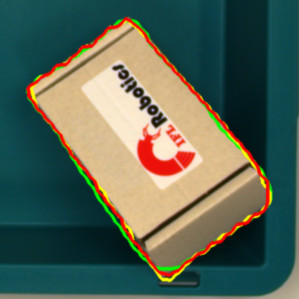}
         \caption{ MC score: 0.968 \\ Confidence: 0.999}
         \label{fig:box_5}
    \end{subfigure}
    \begin{subfigure}[b]{0.17\textwidth}
         \centering
         \includegraphics[width=\textwidth]{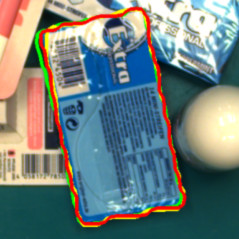}
         \caption{ MC score: 0.968 \\ Confidence: 0.999}
         \label{fig:chewing_gum_with_spray}
    \end{subfigure}
    \begin{subfigure}[b]{0.17\textwidth}
         \centering
         \includegraphics[width=\textwidth]{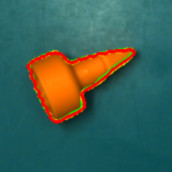}
         \caption{ MC score: 0.957 \\ Confidence: 0.999}
         \label{fig:d_toy_airplane}
    \end{subfigure} \\
    
    \begin{subfigure}[b]{0.17\textwidth}
         \centering
         \includegraphics[width=\textwidth]{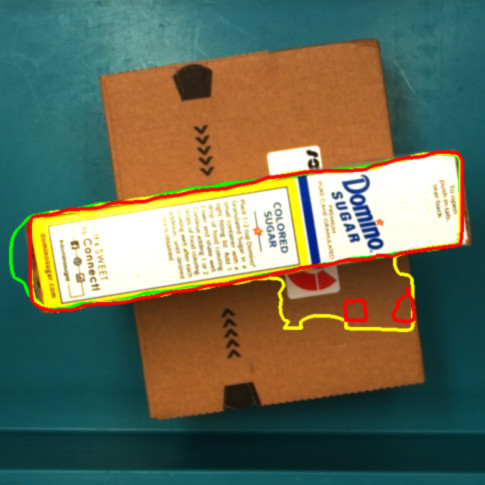}
         \caption{ MC score: 0.888 \\ Confidence: 0.999}
         \label{fig:sugar_box}
    \end{subfigure}
    \begin{subfigure}[b]{0.17\textwidth}
         \centering
         \includegraphics[width=\textwidth]{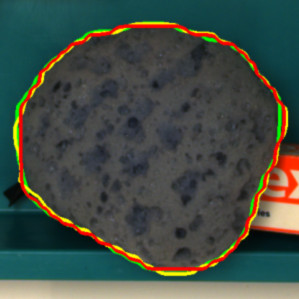}
         \caption{ MC score: 0.972 \\ Confidence: 0.999}
         \label{fig:wash_sponge}
    \end{subfigure}
    \begin{subfigure}[b]{0.17\textwidth}
         \centering
         \includegraphics[width=\textwidth]{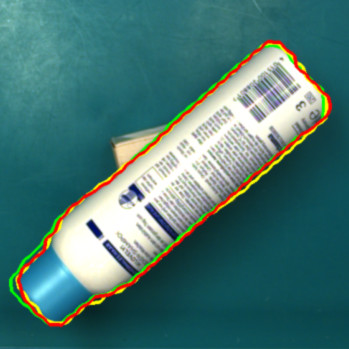}
         \caption{ MC score: 0.953 \\ Confidence: 0.999}
         \label{fig:hairspray}
    \end{subfigure}
    \begin{subfigure}[b]{0.17\textwidth}
         \centering
         \includegraphics[width=\textwidth]{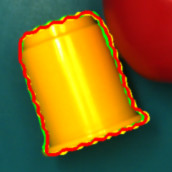}
         \caption{ MC score: 0.965 \\ Confidence: 0.999}
         \label{fig:d_cups}
    \end{subfigure}
    \begin{subfigure}[b]{0.17\textwidth}
         \centering
         \includegraphics[width=\textwidth]{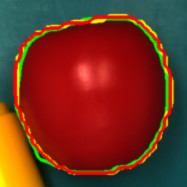}
         \caption{ MC score: 0.961 \\ Confidence: 0.999}
         \label{fig:apple}
    \end{subfigure} \\
    
    \begin{subfigure}[b]{0.17\textwidth}
         \centering
         \includegraphics[width=\textwidth]{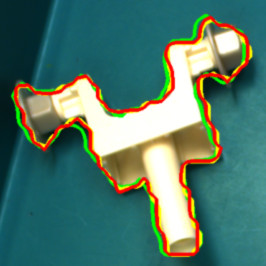}
         \caption{ MC score: 0.908 \\ Confidence: 0.999}
         \label{fig:k_toy_airplane}
    \end{subfigure} 
    \begin{subfigure}[b]{0.17\textwidth}
         \centering
         \includegraphics[width=\textwidth]{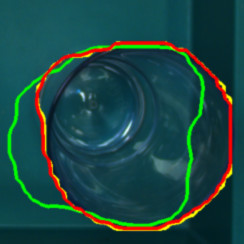}
         \caption{ MC score: 0.854 \\ Confidence: 0.999}
         \label{fig:glass_cup}
    \end{subfigure}
    \begin{subfigure}[b]{0.17\textwidth}
         \centering
         \includegraphics[width=\textwidth]{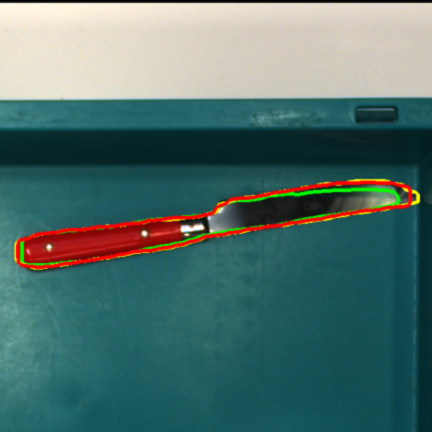}
         \caption{MC score: 0.906 \\ Confidence: 0.999}
         \label{fig:knife}
    \end{subfigure}
    \begin{subfigure}[b]{0.17\textwidth}
         \centering
         \includegraphics[width=\textwidth]{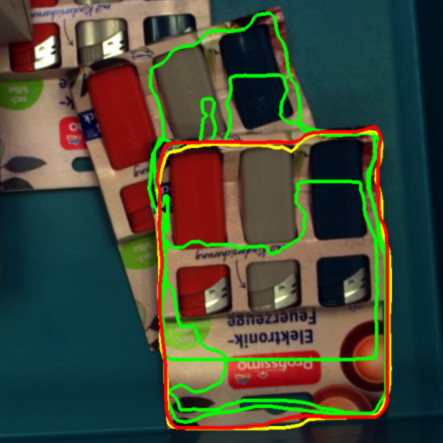}
         \caption{ MC score: 0.749 \\ Confidence: 0.999}
         \label{fig:lighters}
    \end{subfigure}
    \begin{subfigure}[b]{0.17\textwidth}
         \centering
         \includegraphics[width=\textwidth]{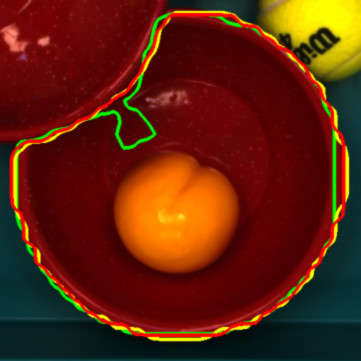}
         \caption{ MC score: 0.749 \\ Confidence: 0.999}
         \label{fig:bowl}
    \end{subfigure} \\
    
    \begin{subfigure}[b]{0.17\textwidth}
         \centering
         \includegraphics[width=\textwidth]{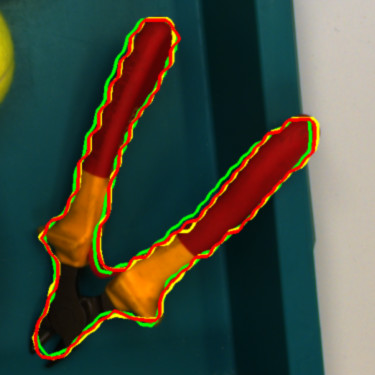}
         \caption{ MC score: 0.909 \\ Confidence: 0.999}
         \label{fig:wire_cutter}
    \end{subfigure}
    \begin{subfigure}[b]{0.17\textwidth}
         \centering
         \includegraphics[width=\textwidth]{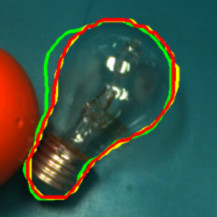}
         \caption{ MC score: 0.935 \\ Confidence: 0.999}
         \label{fig:light_bulb}
    \end{subfigure}
    \begin{subfigure}[b]{0.17\textwidth}
         \centering
         \includegraphics[width=\textwidth]{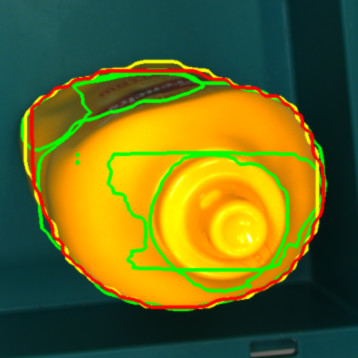}
         \caption{ MC score: 0.751 \\ Confidence: 0.999}
         \label{fig:mustard_bottle}
    \end{subfigure}
    \begin{subfigure}[b]{0.17\textwidth}
         \centering
         \includegraphics[width=\textwidth]{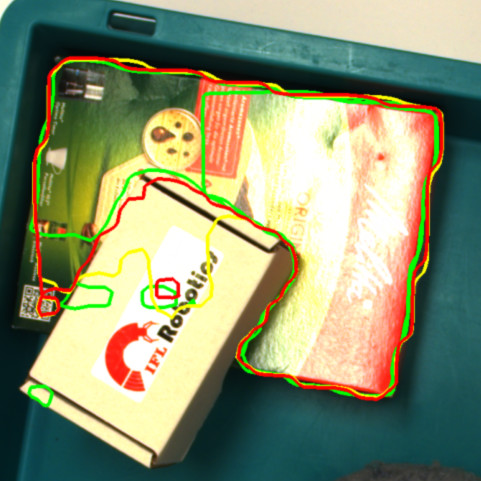}
         \caption{ MC score: 0.794 \\ Confidence: 0.999}
         \label{fig:coffeefilter}
    \end{subfigure}
    \begin{subfigure}[b]{0.17\textwidth}
         \centering
         \includegraphics[width=\textwidth]{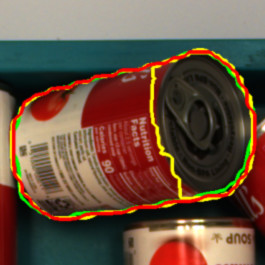}
         \caption{ MC score: 0.772 \\ Confidence: 0.999}
         \label{fig:tomato_soup_can}
    \end{subfigure}

    \caption{Image examples of objects from the real-world test dataset. The MC score shown is calculated at per object level instead of being averaged over the entire object category. We also show output confidence score for comparison. Gate fusion output is marked in red contour. RGB and depth only are marked in yellow and green contours respectively.
    }
    \label{fig:consistency-examples-real}
    
\end{figure*}

In addition, we show randomly selected image results on the real-world test dataset in Figure \ref{fig:consistency-examples-real}.

\begin{table*}[h]
\caption{MC scores (\%) for all object categories in the test-novel set. Item order is ranked based on the combined MC score for mask from low to high.}
\label{table:novel_res_table}
\begin{tabular*}{\textwidth}{rlccccccl}
\toprule
\multirow{2}{*}{Class}    & \multirow{2}{*}{Fig \ref{fig:consistency-examples-novel} ID}    & \multicolumn{3}{c}{MC Score (\%) - Mask} && \multicolumn{3}{c}{MC Score (\%) - Box} \\ \cmidrule{3-5}  \cmidrule{7-9}
                        &  & combined & image  & depth   & &combined & image  & depth  \\
\midrule 
cables in \\transparent bag & (a) & 61.0 & 70.7   & 50.8    && 68.4 & 76.6   & 75.1   \\
eyeglasses                & (b) & 63.7 & 69.6   & 58.0    && 73.6 & 81.1   & 79.0   \\
power drill               & (c) & 63.9 & 68.5   & 73.4    && 68.0 & 77.8   & 74.6   \\
clamp                     & (d) & 68.6 & 67.0   & 74.3    && 75.2 & 83.6   & 80.9   \\
marker small              & (e) & 77.8 & 83.9   & 72.6    && 70.8 & 80.3   & 78.4   \\
mug                       & (f) & 78.6 & 87.1   & 69.2    && 71.6 & 86.6   & 82.8   \\
marker big                & (g) & 79.2 & 86.9   & 84.3    && 67.4 & 77.9   & 77.1   \\
clamp big                 & (h) & 79.4 & 83.6   & 78.2    && 74.6 & 82.7   & 80.7   \\
clamp small               & (i) & 79.7 & 82.7   & 80.0    && 75.5 & 84.0   & 81.1   \\
wineglass                 & (j) & 79.7 & 89.8   & 76.6   & & 72.7 & 81.2   & 79.5   \\
crayons                   & (k) & 81.1 & 83.6   & 78.9   & & 71.7 & 79.8   & 77.7   \\
cables                    & (l) & 85.3 & 89.3   & 86.1   & & 77.1 & 88.5   & 87.8   \\
pear                      & (m) & 91.4 & 94.3   & 93.0   & & 77.6 & 88.8   & 84.7   \\
\bottomrule
\end{tabular*}
\end{table*}

\end{document}